\documentclass[letterpaper, 10 pt, conference]{ieeeconf}

\IEEEoverridecommandlockouts

\usepackage{amsmath}       
\usepackage{amssymb}       
\usepackage{amsfonts}      
\usepackage{mathtools}     
\usepackage{bm}            
\usepackage{bbm}           
\usepackage{xfrac}
\usepackage{times}         
\usepackage{dsfont}        
\usepackage{xspace}        
\usepackage{soul}
\setul{1.5pt}{1pt}

\usepackage{graphicx}      
\usepackage{wrapfig}       
\usepackage{float}         
\usepackage{stfloats}      
\usepackage{capt-of}       
\usepackage[font=footnotesize,labelfont=bf]{caption}      

\usepackage{booktabs}      
\usepackage{multirow}      
\usepackage[flushleft]{threeparttable} 
\usepackage{arydshln}      
\usepackage{listings}
\usepackage[dvipsnames,table,xcdraw]{xcolor} 

\definecolor{mydarkblue}{rgb}{0,0.08,0.45}
\definecolor{mydarkgreen}{RGB}{0, 139, 69}
\definecolor{mygreen2}{RGB}{0, 205, 0}
\definecolor{mybrown}{RGB}{139, 69, 19}
\definecolor{boxblue}{RGB}{79,173,234}
\definecolor{tablepeach}{RGB}{255, 240, 235}
\definecolor{tablepurple}{RGB}{248,235,252}
\definecolor{tableblue}{RGB}{235,241,255}

\usepackage{hyperref}                   
\usepackage[capitalise, nameinlink]{cleveref} 
\usepackage{xurl}                       
\definecolor{citecolor}{HTML}{c03d3e}

\hypersetup{
    colorlinks=true,
    linkcolor=citecolor,
    urlcolor=citecolor,
    citecolor=citecolor,
}
\usepackage[linesnumbered,ruled,vlined]{algorithm2e} 

\usepackage{pifont}  
\usepackage{dsfont}  
\usepackage{lipsum}  
\usepackage{siunitx} 
\usepackage{multicol} 
\let\labelindent\relax
\usepackage{enumitem}

\usepackage[
    style=ieee,
    natbib=true,
    citestyle=numeric-comp,
    doi=false,
    isbn=false,
    url=false
]{biblatex} 
\usepackage{etoolbox} 
\AtBeginBibliography{\footnotesize} 
\newcommand{\ours}{{\color{citecolor}PhysHSI}\xspace}

\newcommand{\ci}[1]{\tiny{\textcolor{gray}{~($\pm #1$)}}}

\title{\LARGE \bf
\ours: Towards a Real-World Generalizable and Natural Humanoid-Scene Interaction System
}

\author{
\authorblockN{
Huayi Wang$^{1,*}$ \quad
Wentao Zhang$^{1,*}$ \quad
Runyi Yu$^{1,2,*}$ \quad
Tao Huang$^{1}$ \quad
Junli Ren$^{1}$ \quad
Feiyu Jia$^{1}$ \quad
Zirui Wang$^{1}$ \\
Xiaojie Niu$^{1}$ \quad
Xiao Chen$^{1}$ \quad
Jiahe Chen$^{1}$ \quad
Qifeng Chen$^{2,\dag}$ \quad
Jingbo Wang$^{1,\dag}$ \quad
Jiangmiao Pang$^{1,\dag}$
}
}

\begin{document}


\twocolumn[{%
\renewcommand\twocolumn[1][]{#1}%
\maketitle
\thispagestyle{empty}
\pagestyle{empty}
\begin{center}
    \centering
    \captionsetup{type=figure}
     \includegraphics[width=1.0\textwidth]{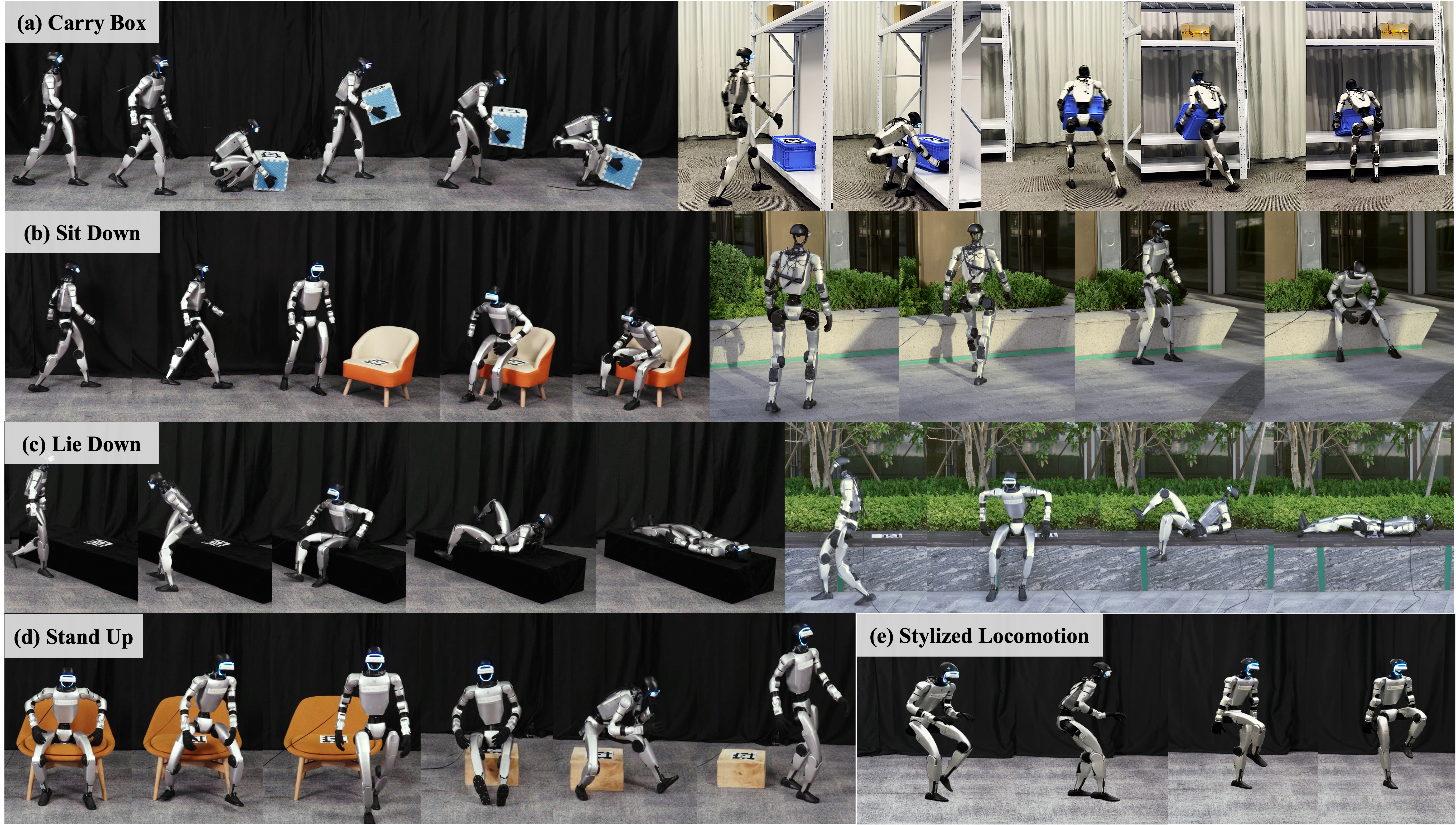}
     \vspace{-0.17in}
    \caption{Our system \ours enables humanoid robots to perform diverse real-world interactions indoors and outdoors with natural behaviors: (a) Carry Box, (b) Sit Down, (c) Lie Down, and (d) Stand Up. \ours can also learn (e) stylized locomotion, such as dinosaur-like walking and high-knee stepping.}
    \label{fig:teaser}
\end{center}
\vspace{-0.0in}
}]

{\footnotesize
\def\thefootnote{} \footnotetext{$^{1}$\,Shanghai AI Laboratory, $^{2}$\,HKUST}
\def\thefootnote{} \footnotetext{*\,Equal Contributions, \dag\,Corresponding Authors}
\def\thefootnote{} \footnotetext{Paper website:
\href{https://why618188.github.io/physhsi}{\texttt{https://why618188.github.io/physhsi}}}
}

\begin{abstract}

Deploying humanoid robots to interact with real-world environments---such as carrying objects or sitting on chairs---requires generalizable, lifelike motions and robust scene perception. Although prior approaches have advanced each capability individually, combining them in a unified system is still an ongoing challenge. In this work, we present  a \ul{phys}ical-world \ul{h}umanoid-\ul{s}cene \ul{i}nteraction system, \ours, that enables humanoids to autonomously perform diverse interaction tasks while maintaining natural and lifelike behaviors. \ours comprises a simulation training pipeline and a real-world deployment system. In simulation, we adopt adversarial motion prior-based policy learning to imitate natural humanoid-scene interaction data across diverse scenarios, achieving both generalization and lifelike behaviors. For real-world deployment, we introduce a coarse-to-fine object localization module that combines LiDAR and camera inputs to provide continuous and robust scene perception. We validate \ours on four representative interactive tasks—box carrying, sitting, lying, and standing up—in both simulation and real-world settings, demonstrating consistently high success rates, strong generalization across diverse task goals, and natural motion patterns.

\end{abstract}

\section{Introduction}

Imagine deploying humanoid robots in everyday environments—carrying boxes into diverse places or sitting naturally on a chair. Building such a humanoid-scene interaction (HSI) system is considered more sophisticated than executing whole-body skills such as standing up~\cite{huang2025learning, he2025learning}, dancing~\cite{cheng2024expressive, ji2024exbody2}, or performing agile motions~\cite{he2025asap, xie2025kungfubot, liao2025beyondmimic}. Beyond these motor capabilities, a real-world HSI system is expected to (1) generalize across diverse interaction scenarios and goals, (2) produce physically plausible and lifelike motions, and (3) incorporate a robust perception module that provides reliable information about surrounding objects and scenes~\cite{gu2025humanoid}. 

Existing approaches fall short of these challenges. While classical model-based methods generate stable motions via motion planning or trajectory optimization for tracking~\cite{ruscelli2020multi, figueroa2020dynamical, sombolestan2023hierarchical, adu2023exploring, liu2025opt2skill}, their high computational cost and strong model assumptions limit generalization to diverse real-world interactions. In contrast, reinforcement learning (RL)-based methods achieve broader generalization by training from diverse simulated experiences. However, learning policies directly from scratch—whether through a single monolithic policy~\cite{zhang2024wococo, liu2024visual, schwarke2023curiosity} or multiple specialized ones~\cite{dao2024sim, zhang2025falcon, lu2024mobile, ben2025homie, xue2025unified}—typically requires heavy reward shaping and state transition design, particularly when natural and lifelike motions are desired. To alleviate the hand-crafted design burden and improve motion realism, methods that imitate motion capture (MoCap) priors have been introduced—these approaches effectively yield physically plausible, human-like motions and have driven progress in physics-based character animation for dynamic interactions~\cite{merel2020catch, hassan2023synthesizing, xie2023hierarchical, gao2024coohoi, xiao2024unified, pan2024synthesizing, pan2025tokenhsi}. However, such approaches largely remain confined to simulation and rely on perfect scene observations, leaving sim-to-real transfer an unexplored obstacle.

In this work, we address these challenges by introducing \ours, a real-world system that enables humanoid robots to autonomously perform HSI skills with natural behaviors across diverse goals and interaction scenarios. The system consists of a simulation training pipeline and a real-world deployment module. In simulation, to learn high-quality humanoid interactions, we first curate retargeted MoCap datasets~\cite{mahmood2019amass, hassan2021stochastic} and augment them with manually annotated object information.  Using these enriched datasets, we then train generalizable HSI policies via reinforcement learning with adversarial motion priors (AMP)~\cite{peng2021amp, escontrela2022adversarial}, leveraging diverse simulation setups to achieve both natural motion and robust generalization. For real-world deployment, where reliable object localization is challenging due to limited fields of view and frequent occlusions, we design a coarse-to-fine perception module that integrates LiDAR-based odometry for long-range directional cues with camera-based object localization for precise pose estimation at close range.

We evaluate \ours on four representative HSI tasks—box carrying and relocation, sitting on chairs, lying on beds, and standing up from chairs~\cite{hassan2023synthesizing}—using Unitree G1 humanoid robots in both simulation and real-world environments. The results show that \ours not only achieves high success rates on these long-horizon tasks but also generalizes effectively across diverse scenarios and task goals. In addition, we demonstrate that \ours produces natural and expressive motions through several learned stylized locomotion behaviors~\cite{mason2022real}. An overview of the system’s real-world performance is provided in Fig.~\ref{fig:teaser}.

In summary, our main contribution is introduing \ours, a real-world HSI system that encompasses: (1) an AMP-based training pipeline in simulation that learns from humanoid interaction data, enabling natural and generalizable motions; (2) a coarse-to-fine real-world object localization module that provides continuous and robust scene perception; and (3) evaluation protocols that comprehensively analyze the system and its components, aiming to guide future research and development in real-world HSI tasks.

\section{Related Works}

\subsection{Humanoid-Scene Interactions}

Many works have studied humanoid-scene interaction (HSI) in physics-based simulations, enabling natural, long-horizon behaviors such as object loco-manipulation~\cite{hassan2023synthesizing, pan2024synthesizing, xiao2024unified, pan2025tokenhsi, wang2024sims}. However, these methods typically rely on idealized task observations and thus face large sim-to-real gaps. For real-world robots, classical approaches often employ model-based motion planning to generate whole-body references for tracking~\cite{ruscelli2020multi, figueroa2020dynamical, sombolestan2023hierarchical, adu2023exploring, liu2025opt2skill}, but these methods exhibit limited generalization in real-world scenarios. In contrast, RL-based methods learn control policies from scratch with strong generalization by carefully designing rewards and state transitions~\cite{zhang2024wococo, schwarke2023curiosity, dao2024sim}. To achieve more natural motion, some works leverage curated motion priors to guide policy learning for tasks such as stair climbing and chair sitting~\cite{allshire2025visual, xue2025leverb}. Building on this line of work, our system learns from motion priors to enable generalizable and natural behaviors for more complex interactions, including box carrying and lying down.

\subsection{Humanoid Motion Imitation}

Humanoid motion imitation seeks to learn lifelike behaviors from human demonstrations, with motion tracking as a central approach. In simulation, physics-based methods achieve expressive whole-body motions by imitating individual reference sequences~\cite{peng2018deepmimic, wang2025skillmimic, yu2025skillmimic} or learning universal tracking~\cite{luo2023universal}. Recent works extend these methods to real-world robots~\cite{cheng2024expressive, ji2024exbody2, he2025asap, liao2025beyondmimic}, but remain reference-dependent and show limited generalization, constraining interactions with diverse scenes. Adversarial Motion Priors (AMP)~\cite{peng2021amp} improve generalization by imitating motion styles and have been widely studied in simulation~\cite{hassan2023synthesizing, pan2024synthesizing, pan2025tokenhsi}. However, real-world applications are limited, with most works using AMP primarily to regularize tracking policies for basic locomotion skills~\cite{xue2025leverb, ma2025styleloco, lin2025hwc, shi2025adversarial}. Building on AMP, our system overcomes these limitations and enables natural behaviors for diverse real-world scene and object interactions.

\subsection{Scene Perception}

Perception is a fundamental component for enabling humanoid robots to interact with real-world scenes and objects. Motion capture (MoCap) systems can provide accurate global information, supporting highly dynamic interactive tasks~\cite{he2024omnih2o, zhang2025unleashing, su2025hitter}. However, MoCap is restricted to laboratory environments with limited workspace. To enable more practical deployment, many studies rely on onboard RGB and depth cameras for scene and object perception~\cite{zhuang2024humanoid, liu2024visual, ma2025learning, fu2024humanplus, lin2025sim, bjorck2025gr00t, qiu2025humanoid}. Yet, these approaches generally confine target objects to a local workspace and often lose sight of them during long-horizon loco-manipulation tasks. Other studies employ LiDAR-Inertial Odometry (LIO)~\cite{xu2022fast,wang2025super} to obtain global information~\cite{wang2025beamdojo, ren2025vb, long2025learning, li2025clone, allshire2025visual}, though interaction accuracy with objects remains limited. In this work, we propose a coarse-to-fine object localization system that relies solely on onboard sensors and provides continuous and robust scene perception.

\section{Simulation Training Pipeline}

\begin{figure*}[t]
    \centering
    \vspace{0.2cm}
    \includegraphics[width=\linewidth]{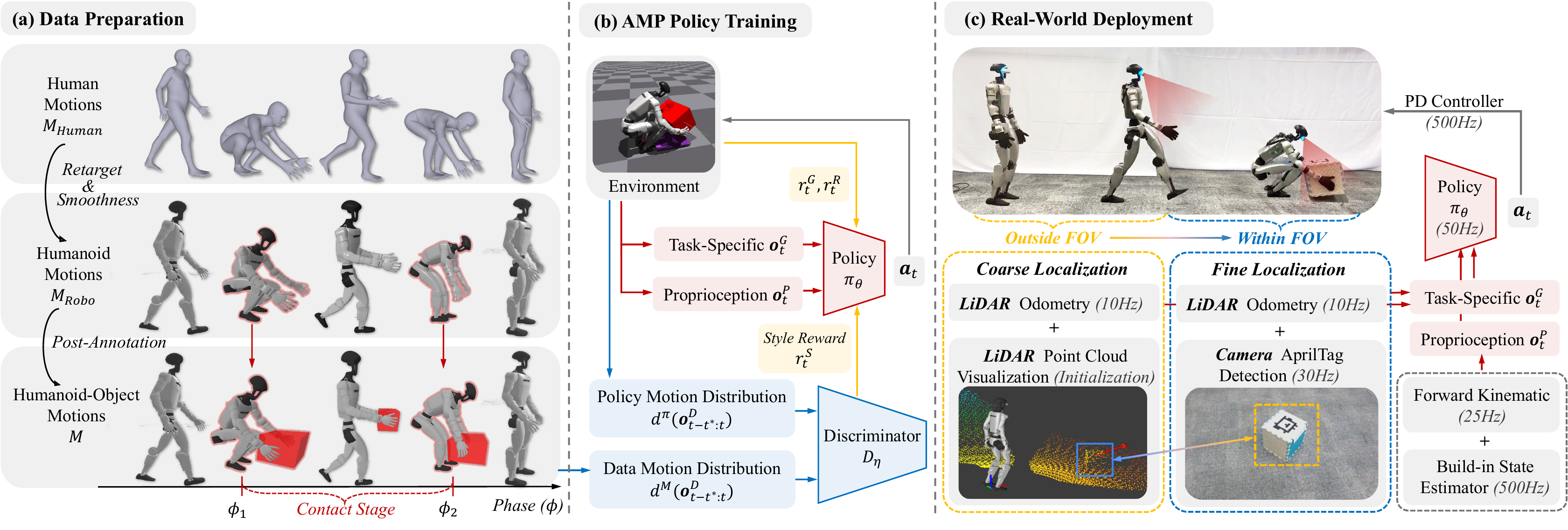}
    \caption{\textbf{Overview of \ours.} 
    (a) \textbf{Dataset Preparation:} Human motions from a MoCap dataset are retargeted to humanoid motions, and objects are annotated by identifying key contact frames.
    (b) \textbf{AMP Policy Training:} A discriminator distinguishes between policy-generated and reference motions to facilitate learning of natural behaviors and task completion.
    (c) \textbf{Real-World Deployment:} The coarse object position is manually specified using LiDAR visualization, and combined with odometry for coarse localization when the object is outside the camera's FOV. Once within view, AprilTag detection combined with odometry is used for fine-grained, automated localization.
    \vspace{-0.2in}
    \vspace{-0.1cm}
    }
    \label{fig:framework}
\end{figure*}

\subsection{Data Preparation}

We begin by preparing humanoid motion data that includes object interactions. While prior works have successfully retargeted human-only motions onto humanoid robots via optimization~\cite{he2025asap, ze2025twist}, generating physically plausible humanoid–object interaction data is more challenging, as it requires maintaining realistic contacts, such as a secure grasp on a box during lifting.

To address this, we adopt a post-annotation strategy for object information. Specifically, we first retarget SMPL motions from the AMASS and SAMP datasets~\cite{mahmood2019amass, hassan2021stochastic} onto the humanoid robot via optimization, applying a smoothing filter to suppress retargeting jitter, yielding a robot-motion-only dataset $M_\text{Robo}$. We then manually annotate key contact frames and infer corresponding object trajectories using a simple rule-based procedure: between pickup ($\phi_1$) and placement ($\phi_2$), the object position $\mathbf{p}^{o_t} \in \mathbb{R}^3$ is set to the midpoint of the hands, with orientation aligned to the robot base; before $\phi_1$ and after $\phi_2$, it remains fixed at the respective key contact frame. This process produces an augmented humanoid motion dataset $M$ with consistent and physically coherent object positions, which is crucial for stage conditioning and reference state initialization (Sec.~\ref{sec:amp}).

\subsection{Adversarial Motion Prior Policy Training}
\label{sec:amp}

We formulate the humanoid-scene interaction (HSI) problem as a reinforcement learning (RL) task. To enable humanoids to interact with objects in a lifelike manner while generalizing across diverse scenarios, we build on the Adversarial Motion Priors (AMP) framework~\cite{peng2021amp}, which has two components: a policy $\pi_\theta$ that generates humanoid actions, and a discriminator $\mathcal{D}$ that distinguishes between policy motions and those in the reference motion dataset.

\subsubsection{Observation and Action Space}

The policy observation $\mathbf{o}^\pi_t$ at each timestep $t$ consists of a 5-step history of proprioception $\mathbf{o}^P_{t-4:t}$ and task-specific observations $\mathbf{o}^G_{t-4:t}$. 

The proprioception $\mathbf{o}^P_t \in \mathbb{R}^{108}$ is defined as
\begin{equation}
   \mathbf{o}^P_t \triangleq \big[ \bm{\omega}_{b_t}, \mathbf{g}_{b_t}, \bm{\theta}_t, \dot{\bm{\theta}}_t, \mathbf{p}^{{ee}}_{b_t}, \mathbf{a}_{t-1} \big], 
\end{equation}
where $b_t$ represent robot base frame $t$, $\bm{\omega}_{b_t} \in \mathbb{R}^3$ is the base angular velocity, $\mathbf{g}_{b_t} \in \mathbb{R}^3$ is the base gravity direction, $\bm{\theta}_t \in \mathbb{R}^{29}$ and $\dot{\bm{\theta}}_t \in \mathbb{R}^{29}$ are joint positions and velocities respectively, $\mathbf{p}^{{ee}}_{b_t} \in \mathbb{R}^{5 \times 3}$ denotes the 3D positions of five end-effectors (left/right hand/foot, and head) in the base frame, and $\mathbf{a}_{t-1}$ is the action taken at the previous timestep.

The task-specific observation $\mathbf{o}^G_t$ varies depending on the task. In general, it consists of three components: (a) the object shape $\mathbf{b}_t \in \mathbb{R}^3$, represented by its bounding box dimensions; (b) the object position $\mathbf{p}^{o_t}_{b_t} \in \mathbb{R}^3$ and orientation $\mathbf{R}^{o_t}_{b_t} \in \mathbb{R}^6$ encoded with a 6D normal-tangent representation~\cite{zhou2019continuity}; and (c) the target goal position $\mathbf{p}^{g_t}_{b_t} \in \mathbb{R}^3$. All quantities are expressed in the robot’s base frame.

The discriminator observation $\mathbf{o}^\mathcal{D}_t \in \mathbb{R}^{57}$ at each timestep consists of privileged information and is defined as
\begin{equation}
    \mathbf{o}^\mathcal{D}_t \triangleq \big[ h_t, \mathbf{v}_{b_t}, \bm{\omega}_{b_t}, \mathbf{g}_{b_t}, \bm{\theta}_t, \mathbf{p}^{{ee}}_{b_t}, \mathbf{p}^{o_t}_{b_t} \big],
\end{equation}
where $h_t \in \mathbb{R}$ denotes the base height, $\mathbf{v}_{b_t} \in \mathbb{R}^3$ is the base linear velocity, $ \bm{\omega}_{b_t}$ is the base angular velocity. Notably, including the object position $\mathbf{p}^{o_t}_{b_t}$ in the discriminator observation is crucial for long-horizon tasks, as it lets the discriminator implicitly condition on task phases—approach, pickup, carry, or place—enhancing policy training guidance.

The action $\mathbf{a}_{t} \in \mathbb{R}^{29}$ from policy $\pi_\theta(\mathbf{o}_t^\pi)$ specifies target joint positions, executed by a PD controller across all 29 humanoid DoFs.

\subsubsection{Reward Terms and Discriminator Learning}

The reward function is defined as the sum of three components: $r_t \triangleq w^G r_t^G + w^R r_t^R + w^S r_t^S$, where $r_t^G$ is the task reward encouraging the humanoid to achieve high-level objectives, $r_t^R$ regularizes excessive joint torques and joint speed, $r_t^S$ is the style reward that encourages the humanoid to imitate behaviors from the reference motion dataset, and $w^{(\cdot)}$ denotes the corresponding coefficients.

The style reward is modeled using the adversarial discriminator $\mathcal{D}$, trained to differentiate between motions produced by the policy and those in the dataset. The discriminator is optimized according to~\cite{peng2021amp}:
\begin{equation}
\begin{aligned}
\arg\min_\mathcal{D} & -\mathbb{E}_{d^{M}(\mathbf{o}^\mathcal{D}_{t:t+t^*})}\left[\log\left(\mathcal{D}(\mathbf{o}^\mathcal{D}_{t:t+t^*})\right)\right] \\
 & -\mathbb{E}_{d^{\pi}(\mathbf{o}^\mathcal{D}_{t:t+t^*})}\left[\log\left(1-\mathcal{D}(\mathbf{o}^\mathcal{D}_{t:t+t^*})\right)\right] \\
 & + w^{\mathrm{gp}} \mathbb{E}_{d^{M}(\mathbf{o}^\mathcal{D}_{t:t+t^*})}\left[\left\|\nabla_{\eta}\mathcal{D}(\eta)\right|_{\eta=(\mathbf{o}^\mathcal{D}_{t:t+t^*})}\right\|^{2}\Big],
\end{aligned}
\end{equation}
where $d^{M}(\mathbf{o}^\mathcal{D}_{t:t+t^*})$ and $d^{\pi}(\mathbf{o}^\mathcal{D}_{t:t+t^*})$ denote the distributions of $(t^*+1)$-frame motion clips from the dataset $M$ and the policy $\pi_\theta$, respectively, and $w^{\mathrm{gp}}$ is a coefficient that regularizes the gradient penalty~\cite{mescheder2018training} in adversarial training. Finally, the style reward for the policy is specified as
\begin{equation}
    r_t^S \triangleq -\log \left( 1-\mathcal{D}(\mathbf{o}^\mathcal{D}_{t-t^*:t}) \right).
\end{equation}

To optimize the policy, we use the proximal policy optimization (PPO)~\cite{schulman2017proximal} to maximize the cumulative discounted reward $\mathbb{E}\left[\sum_{t=1}^{T}\gamma^{t-1}r_{t}\right]$.

\subsubsection{Hybrid Reference State Initialization}

Many HSI tasks are long-horizon, and directly initializing all episodes from the default starting pose makes exploration difficult, since the humanoid rarely experiences critical transitions. To address this, we adopt the reference state initialization (RSI) strategy~\cite{peng2018deepmimic}, which initializes episodes from randomly sampled reference motions along with the corresponding labeled object states, thereby improving exploration efficiency.

This naive RSI strategy, however, risks overfitting to the limited scene configurations in the  demonstrations. We mitigate this limitation in two ways. First, we leverage the compositional nature of task stages: while a motion clip may specify the pickup position of the box, the subsequent goal position does not need to match the data. Thus, we sample an initial phase $\phi \in [0,1]$ from motion data, while randomizing the scene for $(\phi,1]$. Second, a subset of episodes are initialized from the default starting pose with fully randomized scene parameters (e.g., object size, position, and goal position). This hybrid RSI strategy promotes efficient exploration while ensuring generalization.

\subsubsection{Asymmetric Actor-Critic Training}

In real-world, the agent receives only partial observations due to noise and sensing limitations. System constraints further require masking some task observations during training (see Sec.~\ref{sec:sim2real}). To compensate, we adopt the asymmetric actor-critic framework~\cite{pinto2018asymmetric}, where the actor uses inputs $\mathbf{o}^\pi_t$ available at deployment, while the critic observes a richer state $\mathbf{o}^V_t$ (e.g., base velocity and unmasked task observations).

\subsubsection{Motion Constraints}

As rewards accumulate across stages, the agent tends to exploit shortcuts by producing fast, jerky motions, especially later in training, which are unsuitable for deployment. To address this, we assign a small style reward weight $w^S$ early for exploration, and gradually increase it to align with motion data. Additionally, we adopt the L2C2 smoothness regularization~\cite{kobayashi2022l2c2} to enhance smoothness and stability for hardware deployment.

\section{Real-World Deployment System}

To deploy the trained HSI skills in the real world, two key observations must be obtained: the end-effector position $\mathbf{p}_{b_t}^{{ee}}$ and the object pose—position $\mathbf{p}_{b_t}^{o_t}$ and orientation $\mathbf{R}_{b_t}^{o_t}$—in the robot base frame at time $t$. It is easy to get accurate $\mathbf{p}_{b_t}^{{ee}}$ by forward kinematics (FK) with joint encoder information. In contrast, reliable object localization is more challenging, as onboard sensors often suffer from limited fields of view and frequent occlusions—for instance, when the robot starts with no object visible or when the object moves out of view during motion. To overcome these and obtain robust, continuous localization, we design a coarse-to-fine perception system (Sec.~\ref{sec:coarse-to-fine}) that integrates LiDAR and RGB camera inputs. We further adapt the simulation training to align with this perception pipeline (Sec.~\ref{sec:sim2real}) and describe the corresponding hardware setup in Sec.~\ref{sec:hardware}.

\subsection{Coarse-to-Fine Object Localization}
\label{sec:coarse-to-fine}

We represent position and orientation using the transform matrix for clarity. Specifically, 
\begin{equation}
    T_{b_t}^{o_t}=f_{\mathrm{T}}(\mathbf{p}_{b_t}^{o_t},\mathbf{R}_{b_t}^{o_t})\in SE(3)
\end{equation}
denotes the pose of object $o$ at time $t$ in the robot frame $b_t$, where $f_{\mathrm{T}}(\cdot)$ maps position $\mathbf{p}_{b_t}^{o_t}$ and orientation $\mathbf{R}_{b_t}^{o_t}$ to the transform matrix. At initialization, the target object is often outside the camera’s field of view. We therefore assign a coarse initial pose $T_{b_0}^{o_0}$, where the position $\mathbf{p}_{b_0}^{o_0}$ is manually specified using LiDAR point cloud visualization, and the orientation $\mathbf{R}_{b_0}^{o_0}$ is set as default from identity rotation matrix. 

During execution, when the robot is far from the object, we use FAST-LIO~\cite{xu2022fast} to estimate the odometry $T_{b_0}^{b_t}$, i.e., the pose of the current base frame with respect to the initial frame. The object position in the current base frame is then obtained as:
\begin{equation}
\label{eq:propagation}
    \mathbf{p}_{b_t}^{o_t}, \mathbf{R}_{b_t}^{o_t} = f_{\mathrm{T}}^{-1}((T_{b_0}^{b_t})^{-1} T_{b_0}^{o_0}),
\end{equation}
where $f_{\mathrm{T}}^{-1}(\cdot)$ extracts position and orientation from a transformation matrix. This provides a continuous but coarse estimate of the object pose, sufficient to guide the robot toward the target from long range.

For fine-grained localization at close range, AprilTag detection~\cite{wang2016apriltag} is employed to provide accurate object position $\mathbf{p}_{c_t}^{o_t}$ and orientation $\mathbf{R}_{c_t}^{o_t}$ in the camera frame $c_t$. Coarse localization automatically transitions to fine localization upon the tag's first detection. Temporary detection losses (e.g., when the robot turns to sit down) are handled by retaining the last observed object pose $T_{c_{t^\prime}}^{o_{t^\prime}}$ and corresponding FK information $T_{b_{t^\prime}}^{c_{t^\prime}}$, which are then propagated to the current time $t$ using odometry $T_{b_{t^\prime}}^{b_t}$, following the same principle as Eq.~\ref{eq:propagation}:
\begin{equation}
    \mathbf{p}_{b_t}^{o_t}, \mathbf{R}_{b_t}^{o_t} = f_{\mathrm{T}}^{-1}\left((T_{b_{t^\prime}}^{b_t})^{-1} T_{b_{t^\prime}}^{c_{t^\prime}} T_{c_{t^\prime}}^{o_{t^\prime}}\right).
\end{equation}

We further distinguish between static and dynamic objects. For static objects (e.g., chairs), the pose is assumed fixed and updated via the propagation strategy described above, such as when the robot prepares to turn and sit down. For dynamic objects (e.g., boxes), this estimation is valid until grasping; after grasping, if the object leaves the camera view, both position and orientation are masked, and proprioception is relied to complete the task. A simple distance threshold $\epsilon$ defines the grasp phase: if the estimated object distance exceeds $\epsilon$, the object is treated as static; otherwise, it is assumed to move with the robot.

\subsection{Sim-to-Real Transfer}
\label{sec:sim2real}

To better match real-world observations, we apply domain randomization~\cite{tobin2017domain}. Two key strategies are used: (1) adding random offsets, Gaussian noise, and delays to object poses and FK observations; (2) replicating the masking mechanism for dynamic objects during the grasping stage, which is applied when the object is outside the camera’s view, the goal distance is out of range, or the camera angle deviates excessively from vertical. We further adopt standard domain randomization techniques from~\cite{wang2025beamdojo} to enhance robustness and facilitate sim-to-real transfer.

\subsection{Hardware Setup}
\label{sec:hardware}

Our system is built on the Unitree G1 humanoid robot, equipped with a built-in Livox Mid-360 LiDAR and an external Intel RealSense D455 depth camera mounted on the head, with a $86^\circ$ horizontal and $57^\circ$ vertical field of view. Perception modules—including point cloud visualization, Fast-LIO, AprilTag detection, and forward kinematics—together with the learned policy, all run onboard on a Jetson Orin NX, enabling fully portable deployment.

\section{Simulation Experiments}

\begin{table*}[!ht]
\vspace{0.15cm}
\caption{\textbf{Benchmarked Comparison in Simulation.}}
\vspace{-0.1in}
\label{tab:all-results}
\begin{center}
\begin{tabular}{lcccccccc}
\toprule[1.0pt]
\multicolumn{1}{l}{\multirow{2}{*}{}} & \multicolumn{2}{c}{\textit{Carry Box}} & \multicolumn{2}{c}{\textit{Sit Down}} & \multicolumn{2}{c}{\textit{Lie Down}} & \multicolumn{2}{c}{\textit{Stand Up}} \\
\cmidrule[0.7pt](lr){2-3} \cmidrule[0.7pt](lr){4-5} \cmidrule[0.7pt](lr){6-7} \cmidrule[0.7pt](lr){8-9}

\multicolumn{1}{l}{} & 
$R_{\mathrm{succ}} (\%, \uparrow)$ & $S_\mathrm{human} (\uparrow)$ & 
$R_{\mathrm{succ}} (\%, \uparrow)$ & $S_\mathrm{human} (\uparrow)$ & 
$R_{\mathrm{succ}} (\%, \uparrow)$ & $S_\mathrm{human} (\uparrow)$ & 
$R_{\mathrm{succ}} (\%, \uparrow)$ & $S_\mathrm{human} (\uparrow)$\\

\midrule[0.7pt]
\rowcolor[gray]{0.9} \multicolumn{9}{l}{\textbf{\textit{In Distribution Scene}}} \\
\midrule[0.7pt]

RL-Rewards  & 72.92\ci{8.29} & 1.67\ci{0.47} 
           & 83.60\ci{5.98} & 1.50\ci{0.24} 
           & 76.72\ci{9.43} & 0.50\ci{0.00}
           & 93.02\ci{0.71} & 1.50\ci{0.24}\\  [0.4ex]

Tracking-Based  & 11.84\ci{3.16} & \textbf{4.83}\ci{0.24} 
           & 31.46\ci{2.96} & 3.80\ci{0.08} 
           & 19.58\ci{1.02} & 2.23\ci{0.21}
           & 99.00\ci{1.28} & \textbf{4.67}\ci{0.12}\\  [0.4ex]

\ours      & \textbf{91.34}\ci{1.63} & 4.00\ci{0.41} 
           & \textbf{96.28}\ci{0.21} & \textbf{4.80}\ci{0.08} 
           & \textbf{97.86}\ci{0.60} & \textbf{4.80}\ci{0.08}
           & \textbf{99.68}\ci{0.21} & 3.77\ci{0.21}\\

\midrule[0.7pt]

\rowcolor[gray]{0.9} \multicolumn{9}{l}{\textit{\textbf{Full Distribution Scene}}} \\

\midrule[0.7pt]

RL-Rewards  & 63.40\ci{8.63} & 1.17\ci{0.24} 
           & 73.14\ci{4.29} & 3.07\ci{0.09} 
           & 55.76\ci{12.51} & 2.00\ci{1.08}
           & 90.50\ci{2.33} & 1.07\ci{0.09}\\  [0.4ex]

Tracking-Based & 0.02\ci{0.01} & 0.50\ci{0.00} 
           & 1.12\ci{0.51} & 0.50\ci{0.00} 
           & 0.94\ci{0.45} & 1.00\ci{0.41}
           & 35.32\ci{2.51} & 3.27\ci{0.54}\\  [0.4ex]

\ours      & \textbf{84.60}\ci{3.74} & \textbf{3.83}\ci{0.24} 
           & \textbf{91.32}\ci{2.48} & \textbf{4.77}\ci{0.05} 
           & \textbf{81.28}\ci{3.99} & \textbf{4.43}\ci{0.33}
           & \textbf{92.24}\ci{0.75} & \textbf{3.77}\ci{0.52}\\

\bottomrule[1.0pt]
\vspace{-0.9cm}
\end{tabular}
\end{center}
\end{table*}

In this section, we validate the effectiveness of our simulation training pipeline and conduct ablation studies to assess the contribution of each module.

\subsection{Experimental Setup}

We compare \ours to two commonly adopted baselines:
\begin{itemize}[leftmargin=4mm]
    \item \textbf{RL-Rewards}: The humanoid learns HSI tasks from scratch without motion references, using a combination of gait, task, and regularization RL rewards.
    
    \item \textbf{Tracking-Based}: The agent mimics motion references by tracking humanoid and object trajectories provided by the dataset. We use the same dataset as in \ours, which contains roughly 2–5 complete trajectories per task.
\end{itemize}
All training and evaluation environments are implemented in IsaacGym~\cite{makoviychuk2021isaac}. We benchmark methods on four representative HSI tasks:
\emph{\textbf{carry box}} (walk to, lift, carry, and place the box),  
\emph{\textbf{sit down}} (walk to and sit on a chair),  
\emph{\textbf{lie down}} (walk to and lie on a bed), and  
\emph{\textbf{stand up}} (rise from a chair).

For evaluation, we consider two settings: \emph{\textbf{in-distribution scenes}}, which only include scene settings from the dataset, and \emph{\textbf{full-distribution scenes}}, where scenes are uniformly sampled within the task space (objects placed within $[0,5]\,\mathrm{m}$ of the start position; boxes initialized at heights within $[0,0.6]\,\mathrm{m}$ and size dimensions within $[0.2,0.5]\,\mathrm{m}$).

We report two metrics: \textit{\textbf{success rate}} ($R_\mathrm{succ}$) and \textit{\textbf{human-likeness score}} ($S_\mathrm{human}$). $R_\mathrm{succ}$ measures whether the object is correctly placed or the humanoid reaches the desired pose. $S_\mathrm{human}$ is evaluated by Gemini-2.5-Pro~\cite{comanici2025gemini}, which, given task descriptions and experimental trajectories, assigns a human-like score ranging from 0 to 5 for each demonstration clip. For each setting, the mean and standard deviation are computed over five random seeds, each evaluated across 1000 episodes and three demo clips.

\subsection{Overall Performance}

\begin{figure}[t]
    \centering
    \includegraphics[width=\columnwidth]{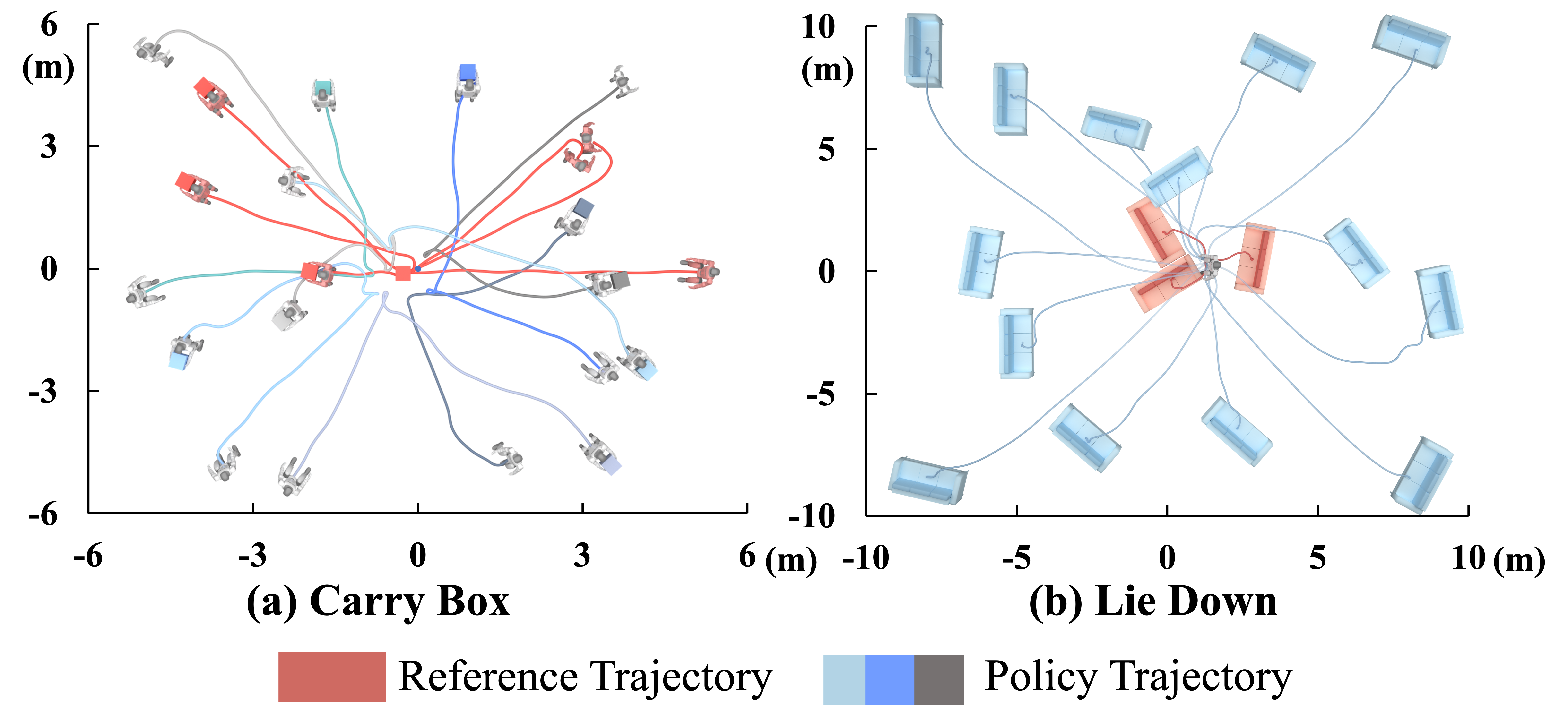}
     \vspace{-0.15in}
    \caption{\textbf{Spatial Generalization.} Root trajectories of the robot are shown for tasks (a) Carry Box and (b) Lie Down. Red trajectories indicate reference data, with others representing sampled policy motions.
    \vspace{-0.18in}}
    \label{fig:generalization}
\end{figure}

\ours achieves high success rates and produces natural, lifelike motions across all tasks, as shown in Table~\ref{tab:all-results}. Key findings are as follows:

\noindent\textbf{Consistently High Success Rates}  
\ours completes all four long-horizon HSI tasks with consistently strong performance. In the more challenging \emph{carry box} task with four subtasks, it reaches an 81.34$\%$ success rate, comparable to the simpler two-step \emph{sit down} task.

\noindent\textbf{Strong Generalization}  
Unlike tracking-based methods that mimic reference trajectories frame by frame, \ours leverages AMP frameworks to enable flexible motion recombination, requiring only style alignment with motion priors. This enables comparable success rates even in full-distribution scenes, whereas tracking-based methods almost completely fail (near 0 success) due to the limited scale of reference data. Fig.~\ref{fig:generalization} shows partially successful trajectories, highlighting the strong generalization learned from only a few references.

\noindent\textbf{Lifelike Motion Patterns}  
\ours attains significantly higher $S_\mathrm{human}$ than RL reward–based methods. By training the policy and discriminator in competition, our approach effectively distinguishes between dataset motions and policy-generated motions, leading to natural behaviors. In contrast, RL reward methods require carefully hand-crafted gait and regularization terms, which are difficult to design and less effective for long-horizon tasks.

\subsection{Ablation Analysis}

We conduct ablation studies on the data processing, RSI strategy and mask strategy. The main results are presented in Table~\ref{tab:ablation-results}, with the following key observations:

\textbf{Data quality and object annotation are critical for natural motion and task completion.}  
Training without smoothed motion data (w/o Smoothness) produces unnatural behaviors, as the policy—guided by the discriminator—may exploit artifacts such as jittering end-effectors or abrupt motion shifts. Removing object annotations (w/o Object) increases failure rates, since object states in AMP observations are essential for learning stage transitions and motion styles. For instance, when the object is distant, the discriminator drives the humanoid to walk toward it, while during carrying, it keeps the box centered between the hands.

\textbf{Hybrid RSI is crucial for generalization and efficiency.}
We compare our hybrid RSI with two alternatives: no RSI (w/o RSI) and naive RSI, where all episodes are initialized from reference states fixed to dataset settings. Naive RSI performs worse than no RSI, demonstrating poor generalization and low training efficiency due to the limited diversity of observed scenes. In contrast, hybrid RSI significantly improves both generalization and sample efficiency.

\textbf{Mask processing has a limited impact on overall performance.}
Although the masking strategy introduced in Sec.~\ref{sec:sim2real} slightly slows training compared to using complete object states (w/o Obs Mask), which represent an upper bound on performance, it only minimally affects the final policy success rate in the two ablation cases.

\vspace{0.2cm}
\setlength{\tabcolsep}{4pt}
\begin{table}[t]
\vspace{0.15cm}
\caption{\textbf{Ablation Experiments.}}
\label{tab:ablation-results}
\vspace{-0.1in}
\begin{center}
\resizebox{0.485\textwidth}{!}{
\begin{tabular}{lcccc}
\toprule[1.0pt]
\multicolumn{1}{l}{\multirow{2}{*}{}} & \multicolumn{2}{c}{\textit{Carry Box}} & \multicolumn{2}{c}{\textit{Sit Down}} \\
\cmidrule[0.7pt](lr){2-3} \cmidrule[0.7pt](lr){4-5} 

\multicolumn{1}{l}{} & 
$R_{\mathrm{succ}} (\%, \uparrow)$ & $S_\mathrm{human} (\uparrow)$ & 
$R_{\mathrm{succ}} (\%, \uparrow)$ & $S_\mathrm{human} (\uparrow)$ \\

\midrule[0.7pt]
\rowcolor[gray]{0.9} \multicolumn{5}{l}{\textbf{\textit{Ablation on Data Processing}}} \\
\midrule[0.7pt]

w/o Smoothness
& 63.28\ci{11.72} & 2.33\ci{1.03} 
& 87.24\ci{2.19} & 1.33\ci{0.23} \\[0.4ex]

w/o Object
& 55.42\ci{8.17} & 2.60\ci{0.57} 
& 72.36\ci{6.71} & 3.50\ci{0.70} \\[0.4ex]

\ours 
& \textbf{79.34}\ci{4.71} & \textbf{3.83}\ci{0.24} 
& \textbf{91.32}\ci{2.48} & \textbf{4.77}\ci{0.05} \\

\midrule[0.7pt]

\rowcolor[gray]{0.9} \multicolumn{5}{l}{\textit{\textbf{Ablation on RSI Strategy}}} \\

\midrule[0.7pt]

w/o RSI  
& 41.24\ci{6.92} & 2.50\ci{1.63} 
& 78.24\ci{3.91} & 4.50\ci{0.00} \\  [0.4ex]

Naive RSI  
& 5.70\ci{2.38} & 0.50\ci{0.0} 
& 18.70\ci{5.33} & 1.83\ci{0.62} \\  [0.4ex]

Hybrid RSI   
& \textbf{79.34}\ci{4.71} & \textbf{3.83}\ci{0.24} 
& \textbf{91.32}\ci{2.48} & \textbf{4.77}\ci{0.05} \\

\midrule[0.7pt]

\rowcolor[gray]{0.9} \multicolumn{5}{l}{\textit{\textbf{Ablation on Mask Strategy} (for dynamic objects)}} \\

\midrule[0.7pt]

w/o Obs Mask  
& \textbf{85.90}\ci{2.90} & \textbf{4.30}\ci{0.14} 
& / & / \\  [0.4ex]

\ours
& 79.34\ci{4.71} & 3.83\ci{0.24} 
& \textbf{91.32}\ci{2.48} & \textbf{4.77}\ci{0.05} \\

\bottomrule[1.0pt]
\vspace{-0.9cm}
\end{tabular}
}
\end{center}
\end{table}

\section{Real-World Experiments}

In this section, we evaluate the overall system performance in real-world scenarios and assess the effectiveness of our proposed coarse-to-fine object localization module.

\subsection{Overall Performance}

\begin{figure*}[t]
    \centering
    \vspace{0.15cm}
    \includegraphics[width=\linewidth]{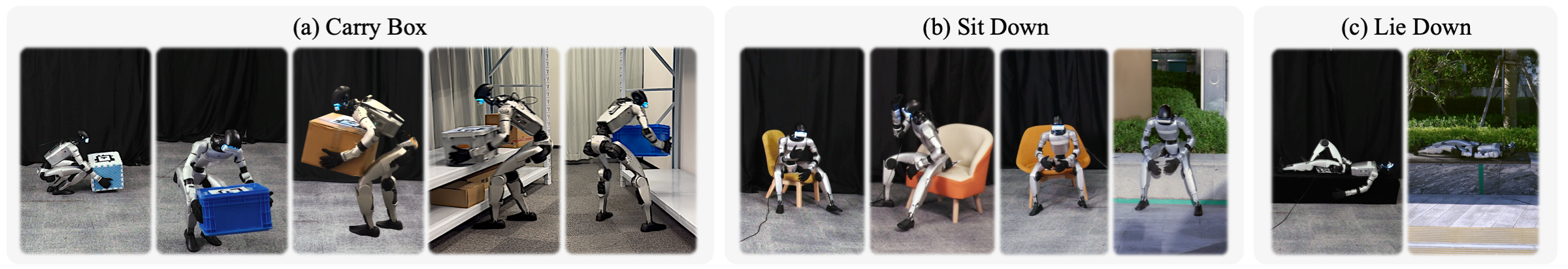}
    \caption{\textbf{Real-World Generalization.} \ours generalizes to diverse real-world scenes, (a) handling boxes of varying shapes, weights, and heights, and (b) sitting or (c) lying on chairs and beds of different heights, both indoors and outdoors.}
    \vspace{-0.1in}
    \vspace{-0.05cm}
    \label{fig:real_generalization}
\end{figure*}

As shown in Fig.~\ref{fig:teaser}, \ours achieves zero-shot transfer and successfully completes all four HSI tasks in real-world settings. We further evaluate success rate $R_\mathrm{succ}$, finish precision $R_\mathrm{precision}$, execution time $T_\mathrm{exec}$, and maximum movement range $M_\mathrm{range}$ with 10 trials per task, as reported in Table~\ref{tab:real_results}. Our key findings are summarized below:

\begin{itemize}[leftmargin=4mm]
    \item \ours achieves competitive success rates with high precision in real-world deployments across all four tasks, showing particularly strong performance on \emph{lie down} and \emph{sit down}. For the more challenging \emph{carry box} task, the system attains an 8/10 success rate for lifting and 6/10 for the full sequence, with placement errors under 20\,$\mathrm{cm}$.
    
    \item \ours generalizes effectively across variations in spatial layout and object properties, handling locomotion over distances up to 5.7\,$\mathrm{m}$ with diverse box dimensions, heights, and weights. Representative examples of varied scene configurations are shown in Fig.~\ref{fig:real_generalization}.
    
    \item Compared to reward-tuned RL policies, \ours generates more natural, human-like motions. Our policy inherits the catwalk-style locomotion present in AMASS data, while the framework also supports stylized motion learning. As shown in Fig.~\ref{fig:teaser}(e), the system can produce diverse locomotion styles, such as dinosaur-like walking or high-knee stepping.

    \item \ours can be deployed outdoors using only onboard sensing and computation (Fig.~\ref{fig:teaser}(a)-(c)). This highlights the portability of our system compared to MoCap-based deployments that rely on external infrastructure.
\end{itemize}

\subsection{Object Localization Module Analysis}

\begin{table}[t]
\vspace{0.15cm}
\caption{\textbf{Real-World Experiments.} Success rates for the pick-up stage and the full sequence are separately reported for the \textit{Carry Box} task.
\vspace{-0.25cm}
}
\label{tab:real_results}
\begin{center}
\begin{tabular}{l>{\centering\arraybackslash}p{1.4cm}ccc}
\toprule[1.0pt]
\textbf{Tasks} & $R_\mathrm{succ}$ & $R_\mathrm{precision}\,(\mathrm{m})$ & $T_\mathrm{exec}\,(\mathrm{s})$ & $M_\mathrm{range}\,(\mathrm{m})$ \\
\midrule[0.7pt]

\textit{Carry Box}
& 8$/$10,\ 6$/$10 & 0.19\ci{0.10} 
& 10.5\ci{2.8} & 5.69 \\[0.4ex]

\textit{Sit Down}
& 9$/$10 & 0.07\ci{0.03} 
& 6.2\ci{1.3} & 4.14 \\[0.4ex]

\textit{Lie Down}
& 8$/$10 & 0.16\ci{0.07} 
& 6.7\ci{1.0} & 3.76 \\[0.4ex]

\textit{Stand Up}
& 8$/$10 & $/$
& 2.3\ci{0.4} & 1.74 \\

\bottomrule[1.0pt]
\vspace{-0.9cm}
\end{tabular}
\end{center}
\end{table}

\begin{figure}[t]
    \centering
    \includegraphics[width=1.0\linewidth]{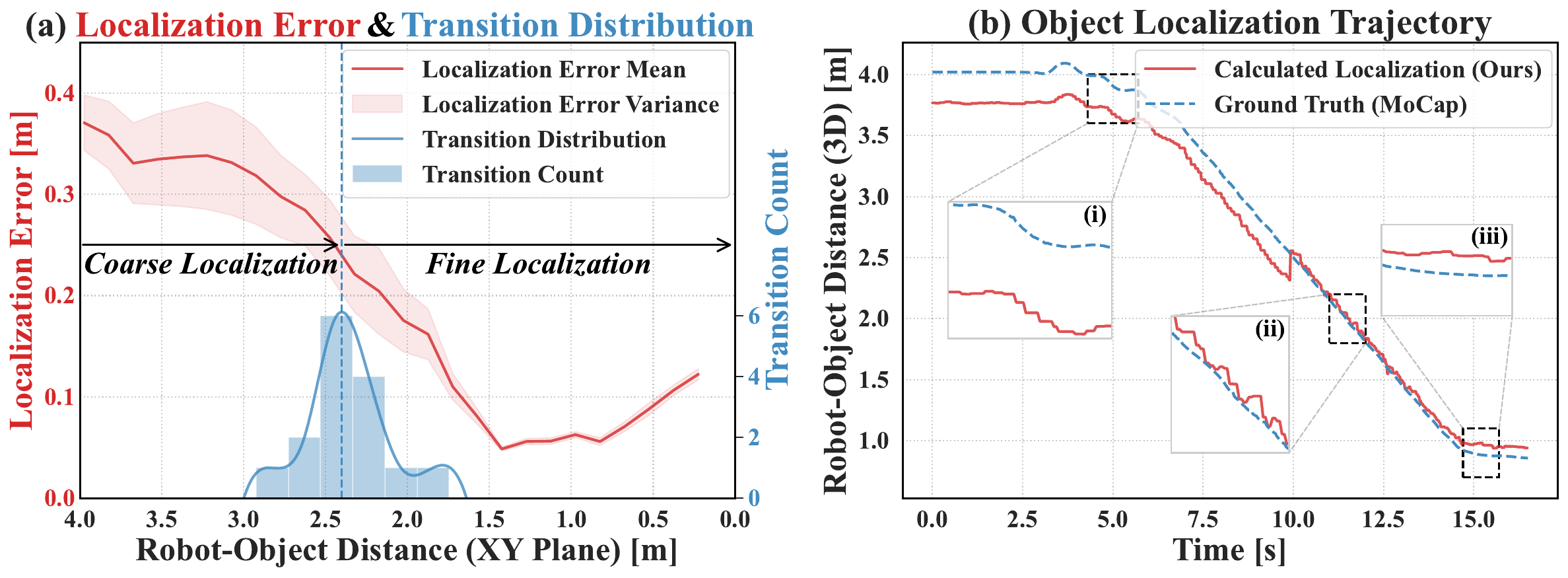}
    \vspace{-0.22in}
    \caption{\textbf{Real-World Localization System Analysis.} (a) Localization error versus robot–object distance, with coarse-to-fine transition statistics and distribution. 
    (b) A representative object localization trajectory, highlighting three stages: (i) coarse localization, (ii) fine localization, and (iii) grasp.
    \vspace{-0.18in}
    }
    \label{fig:real_system_analysis}
\end{figure}

To evaluate the effectiveness of our object localization module, we conducted 17 real-world HSI trials, 15 of which were successful. For each trial, we recorded the object trajectories estimated by our module and compared them against ground-truth trajectories obtained from a MoCap system. We also measured the robot-object distance at the coarse-to-fine transition point. As shown in Fig.~\ref{fig:real_system_analysis}(a), localization error is relatively large (0.35\,$\mathrm{m}$) when the robot is far from the object. Once within 2.4\,$\mathrm{m}$, AprilTag detection activates, switching to fine localization with an average error of 0.05\,$\mathrm{m}$. These results demonstrate the effectiveness of our design: the coarse stage provides reliable directional guidance at long range, while the fine stage yields accurate positions at close range. The overall success rate of 15/17 further confirms robustness. Two failures occurred due to coarse guidance deviating too far (preventing tag from entering FOV) and a system crash.

To further analyze error sources across stages, we examine one successful trajectory by comparing the estimated localization with the ground-truth trajectory (Fig.~\ref{fig:real_system_analysis}(b)). Three stages are observed:  
(i) \textbf{Coarse stage}: errors mainly arise from manually specified goal points in LiDAR point-cloud visualization, deviating ~0.3\,$\mathrm{m}$ from the exact position. Despite this, the estimated and ground-truth trajectories show consistent trends, which is sufficient for guidance.  
(ii) \textbf{Fine stage}: errors stem from odometry drift and AprilTag noise, but remain small, with trajectories closely aligned.
(iii) \textbf{Grasping stage}: at close range, errors are dominated by AprilTag noise and are more pronounced due to rapid manipulative motions compared to smoother locomotion.

\subsection{System Limitation Analysis}

\begin{table*}[t]
\centering
\caption{\textbf{Limitation Test for \textit{Carry Box} Task.}}
\vspace{-0.05in}
\setlength{\tabcolsep}{5pt}
 \begin{tabular}{lc c cccc c ccccc c cccc c} 
 \toprule[1.0pt]
  \multirow{2}{*}{\textbf{Test Condition}} & & \multicolumn{4}{c}{Box Height} &  &\multicolumn{5}{c}{Box Weight} & & \multicolumn{5}{c}{Maximum Box Size} \\
  
   \cmidrule[0.7pt]{3-6} \cmidrule[0.7pt]{8-12} \cmidrule[0.7pt]{14-17} 
    & 
    & 0\,$\mathrm{cm}$ & 20\,$\mathrm{cm}$ & 40\,$\mathrm{cm}$ & 60\,$\mathrm{cm}$
    & 
    & 0.6\,$\mathrm{kg}$ & 1.2\,$\mathrm{kg}$ & 2.3\,$\mathrm{kg}$ & 3.6\,$\mathrm{kg}$ & 4.5\,$\mathrm{kg}$
    &
    & 20\,$\mathrm{cm}$ & 30\,$\mathrm{cm}$ & 40\,$\mathrm{cm}$ & 45\,$\mathrm{cm}$
    & \\
 \midrule[0.7pt]
 
 $R_{\mathrm{succ}}(\uparrow)$ & & 
 2$/$3 & 3$/$3 & 3$/$3 & 1$/$3 & &
 2$/$3 & 3$/$3 & 2$/$3 & 1$/$3 & 0$/$3 & &
 2$/$3 & 3$/$3 & 2$/$3 & 3$/$3 & \\
\bottomrule[1.0pt]
\end{tabular}
\label{tab:limitation_test}
\vspace{-0.12in}
\end{table*}

We analyze the limitations of our system on the \emph{carry box} task. We evaluate different carrying heights, box masses, and shapes, each with three trials, and report the success rates in Table~\ref{tab:limitation_test}. We find that the humanoid can stably carry boxes at heights in $[0, 60]~\mathrm{cm}$, with weights in $[0.6, 3.6]~\mathrm{kg}$, and maximum sizes up to $[20, 45]~\mathrm{cm}$. Carrying higher boxes exceeds the robot's vertical FOV even when stationary, while heavier or wider boxes cannot be handled due to the limited reach of the rubber hand and arm length.

Beyond these findings, we identify several broader limitations that highlight challenges in advancing HSI capabilities:

\noindent\textbf{Hardware Constraints.} The current system relies on the Unitree G1’s rubber hand for clamping, which restricts manipulation of larger or heavier boxes. Excessive weight may also cause motor overheating and potential hardware failures during execution.

\noindent\textbf{Large-Scale High-Quality HSI Data.} In this work, we post-annotate objects in retargeted humanoid motion data and select a small subset of high-quality samples for training. However, this manual process does not scale well when large scale of high-quality HSI data are required.  

\noindent\textbf{Automated Perception Module.} Our current object localization relies on a modular real-world system, which introduces complexity and potential fragility. Developing a more automated perception module, for example with active perception that enables autonomous exploration, could improve robustness and simplify deployment.

\section{Conclusions}

We presented \ours, a real-world system for generalizable and natural humanoid-scene interaction, combining an effective simulation training pipeline with a robust deployment module. \ours successfully performs tasks such as \emph{carry box} and \emph{lie down} in real-world scenarios with high success rates, strong spatial and object-level generalization, and natural motion behaviors. Moreover, with only a manually specified coarse object initialization and a single fiducial tag, our system can autonomously complete tasks even in outdoor environments, demonstrating its portability. This work represents an initial exploration of real-world HSI tasks and paves the way for more advanced object- and scene-interaction capabilities in practical applications.

\section*{Acknowledgements}

This work is funded in part by the National Key R\&D
Program of China (2022ZD0160201), and Shanghai Artificial Intelligence Laboratory. We thank Liang Pan for advice on the implementation of RSI. We thank Shunlin Lu for help with the process of motion data. We thank Jianhui Liu, Tai Wang, Qingwei Ben and Junfeng Long for valuable discussions and advice on the object localization module. We thank Chenhui Li and Intelligent Photonics and Electronics Center at Shanghai AI Lab for help with the MoCap system and SLAM devices. We thank Weixiang Zhong and Yinhuai Wang for assistance with the real-world experiments. We thank Unitree and the Hardware Team of the Embodied AI Center at Shanghai AI Lab for help with hardware issues and the Unitree G1 humanoid robot.

\vspace{0.2cm}
\printbibliography

\clearpage
\section*{Appendix}

\subsection{Tasks}

In this section, we provide detailed definitions of each task, including the composition of the reference motion dataset $M$, the task-specific observation $\mathrm{o}_t^G$, and the task reward $r_t^G$.

\subsubsection{\textbf{Carry Box}}
The humanoid starts from a random position and is tasked with approaching and moving a box from a randomly initialized 3D location to a target 3D location. In simulation, two thin platforms are used to support the box, since both the initial and target heights are randomly generated.

\noindent \textbf{Reference Motion Dataset} The motion dataset for \textit{Carry Box} consists of two parts. The first part, \textit{Loco}, includes 11 motion sequences from the AMASS~\cite{mahmood2019amass} dataset, covering basic locomotion behaviors such as standing, walking, and turning on flat ground. The second part, \textit{Carry}, includes 3 sequences from the AMASS dataset and 2 video-based motion sequences, which were retargeted to SMPL motions using GVHMR~\cite{shen2024world} and subsequently refined by manually correcting certain joints to ensure better physical feasibility. For convenience in RSI, this dataset is further divided into three subsets: \textit{pickUp}, \textit{carryWith}, and \textit{putDown}.

\noindent \textbf{Task Observations} The task-specific observation $\mathbf{o}_t^G \in \mathbb{R}^{15}$ comprises the following properties of the target box:
\begin{itemize}[leftmargin=4mm]
\item Box shape $\mathbf{b}_t \in \mathbb{R}^3$
\item Box position $\mathbf{p}_{b_t}^{o_t} \in \mathbb{R}^3$
\item Box rotation $\mathbf{R}_{b_t}^{o_t} \in \mathbb{R}^6$
\item Goal location of the box $\mathbf{p}_{b_t}^{g_t} \in \mathbb{R}^3$
\end{itemize}

\noindent \textbf{Task Rewards} We implement the multi-stage task reward function similar to TokenHSI~\cite{pan2025tokenhsi}. The first stage aims to encourage the robot to walk toward the initial box:
\begin{equation}
r_t^{loco} = \left\{
    \begin{aligned}
    &1.5, \  \left \| \mathbf{p}^{o_t}_{xy} - \mathbf{p}^{b_t}_{xy} \right \| < 0.7 \\
    &1.0 \exp \big (-5.0 \left \| 0.85 - \mathbf{d}_t^* \cdot \dot{\mathbf{p}}^{b_t}_{xy} \right \|^2 \big )\,+ \\
    &\quad 0.5 \exp \big(-0.75 \left \|\Delta\theta(\mathbf{d}_t^*, \mathbf{d}_{b_t}) \right \| \big),\ \text{otherwise}
    \end{aligned}
    \right.
\end{equation}
where $\mathbf{p}^{o_t}_{xy}$ and $\mathbf{p}^{b_t}_{xy}$ denote the 2D positions of the object and the robot base in the world frame, respectively. $\mathbf{d}_t^*$ is a horizontal unit vector pointing from $\mathbf{p}^{b_t}_{xy}$ to $\mathbf{p}^{o_t}_{xy}$, $\mathbf{d}_{b_t}$ is the 2D horizontal unit vector of the robot base orientation in the world frame, and $a \cdot b$ represents the vector dot product. $\Delta\theta(\mathbf{d}_t^*, \mathbf{d}_{b_t})$ is the yaw error between target heading and root heading, defined as
\begin{equation}
\Delta\theta(\mathbf{a}, \mathbf{b}) = \arctan2(\mathbf{a}_y, \mathbf{a}_x) - \arctan2(\mathbf{b}_y, \mathbf{b}_x),
\end{equation}
where $\mathbf{a}$ and $\mathbf{b}$ are 2D horizontal vectors. The second stage is to encourage the robot to pick up and move the box to its target location, which is defined:
\begin{equation}
r_t^{carry} = \left\{
    \begin{aligned}
    &0.0, \ \left \| \mathbf{p}^{o_t}_{xy} - \mathbf{p}^{b_t}_{xy} \right \| > 0.7\\
    &2.2, \ \left \| \mathbf{p}^{b_t}_{xy} - \mathbf{p}^{g_t}_{xy} \right \| < 0.7 \\
    &1.0 \exp \big (-5.0 \left \| 0.85 - \mathbf{d}_t^\# \cdot \dot{\mathbf{p}}^{b_t}_{xy} \right \|^2 \big )\,+ \\
    &\quad 0.5 \exp \big(-0.75 \left \|\Delta\theta(\mathbf{d}_t^\#, \mathbf{d}_{b_t}) \right \| \big)\,+  \\
    &\quad 0.7 \exp \big( -3.0 \left \| \mathbf{p}^{o_t} - \mathbf{p}^{hand_t} \right \|^2 \big),\ \text{otherwise}
    \end{aligned}
    \right.
\end{equation}
where $\mathbf{p}^{g_t}_{xy}$ denotes the 2D positions of the goal in the world frame, $\mathbf{d}_t^\#$ is a horizontal unit vector pointing from $\mathbf{p}^{b_t}_{xy}$ to $\mathbf{p}^{g_t}_{xy}$, and $\mathbf{p}^{hand_t}$ denotes the mean 3D coordinates of the robot’s two hands. The third term of $r_t^{carry}$ encourages the robot to pick up the box using its hands. 
To further reinforce this behavior, we additionally reward the lifting height during the pickup stage, defined as:
\begin{equation}
r_t^{pick} = \left\{
    \begin{aligned}
    &0.0, \  \left \| \mathbf{p}^{o_t}_{xy} - \mathbf{p}^{b_t}_{xy} \right \| > 0.7 \\
    &2.0, \  \left \| \mathbf{p}^{b_t}_{xy} - \mathbf{p}^{g_t}_{xy} \right \| < 0.7 \  \text{or} \  \mathbf{p}^{o_t}_{z} > 0.75\\
    &2.0 \exp \big (-3.0 \left \| 0.75 - \mathbf{p}^{o_t}_{z} \right \| \big ),\ \text{otherwise}
    \end{aligned}
    \right.
\end{equation}
where $\mathbf{p}^{o_t}_{z}$ denotes the height of the box in the world frame. Additionally, we further design a reward function $r_t^{put}$ to encourage the robot to accurately place the box at the target location:
\begin{equation}
r_t^{put} = \left\{
    \begin{aligned}
    &0.0, \  \left \| \mathbf{p}^{b_t}_{xy} - \mathbf{p}^{g_t}_{xy} \right \| > 0.7\\
    &2.0, \  \left \| \mathbf{p}^{o_t} - \mathbf{p}^{g_t} \right \| < 0.05 \\
    &1.0 \exp \big (-10.0 \left \| \mathbf{p}^{o_t} - \mathbf{p}^{g_t} \right \| \big )\,+ \\
    &\quad 1.0 \exp \big(-3.0 \left(\mathbf{p}^{o_t}_z - \mathbf{p}^{g_t}_z \right) \big),\ \text{otherwise}
    \end{aligned}
    \right.
\end{equation}
where $\mathbf{p}^{g_t}_{z}$ denotes the height of the goal in the world frame. Therefore, the total task reward function for \textit{Carry Box} be formulated as:
\begin{equation}
r_t^{G\_carryBox} = r_t^{loco} + r_t^{carry} + r_t^{pick} + r_t^{put}.
\end{equation}

\noindent \textbf{Scene Randomization} We randomize the task scene along the following four dimensions:
\begin{itemize}[leftmargin=4mm]
\item The 2D position of the box and the target relative to the robot’s initial position is uniformly sampled from $[-4.0, 4.0]\,\mathrm{m}$ relative to the robot’s initial base position.
\item The height position of both the box and the target is uniformly sampled from $[0.0, 0.6]\,\mathrm{m}$ above the ground.
\item The box size is randomized, with the width uniformly sampled from $[0.2, 0.5]\,\mathrm{m}$ and the height uniformly sampled from $[0.15, 0.35]\,\mathrm{m}$.
\item The box density is uniformly sampled from $[10, 100]\,\mathrm{kg/m^3}$.
\end{itemize}

\noindent \textbf{Observation Mask Strategy} As discussed in Sec.~\ref{sec:sim2real}, we align the simulation training with the real-world deployment by masking object observations $\mathbf{p}_{b_t}^{o_t}$ and $\mathbf{R}_{b_t}^{o_t}$ when the box is \textit{out of view} during the grasping stage of dynamic objects. We define an object as \textit{out of view} in simulation if it satisfies one of the following conditions:
\begin{itemize}[leftmargin=4mm]
\item Facing condition: The surface normal of the box must face toward the camera, i.e., the angle between the viewing direction and the surface normal is within $(60^\circ+\Delta)$, where the offset $\Delta$ is uniformly sampled from $[-10^\circ, 10^\circ]$.
\item FOV condition: All tag positions of the box must lie within the camera’s field-of-view (FOV), constrained by both the horizontal and vertical FOV angles.
\item Distance condition: The mean tag position must lie within $2.5\,\mathrm{m}$ range from the camera.
\end{itemize}

\subsubsection{\textbf{Sit Down and Lie Down}}
The humanoid starts from a random position and is tasked with approaching a fixed chair or bed surface to perform a sitting or lying action. In the simulation, a thin platform and a box are used to support sitting and lying behaviors, respectively.

\noindent \textbf{Reference Motion Dataset} For locomotion, we use the same \textit{Loco} dataset as in the \textit{Carry Box} task. For the \textit{Sit Down} task, we additionally select four sitting sequences from SAMP~\cite{hassan2021stochastic}, and for the \textit{Lie Down} task, we additionally select six lying sequences from SAMP.

\noindent \textbf{Task Observations} The task-specific observation $\mathbf{o}_t^G \in \mathbb{R}^{9}$ comprises the following properties of the target chair/bed:
\begin{itemize}[leftmargin=4mm]
\item Chair/bed position $\mathbf{p}_{b_t}^{o_t} \in \mathbb{R}^3$
\item Chair/bed rotation $\mathbf{R}_{b_t}^{o_t} \in \mathbb{R}^6$
\end{itemize}

\noindent \textbf{Task Rewards} For the locomotion stage, both tasks share the same locomotion reward $r_t^{loco}$ as in the \textit{Carry Box} task. The second stage encourages the robot to sit down on the chair or bed, with the sitting reward defined as:
\begin{equation}
r_t^{sit} = \left\{
    \begin{aligned}
    &0.0, \  \left \| \mathbf{p}^{o_t}_{xy} - \mathbf{p}^{b_t}_{xy} \right \| > 0.7 \\
    &1.0 \exp \big (-3.0 \left \| \mathbf{p}^{o_t} - \mathbf{p}^{b_t} \right \| \big )\,+ \\
    &\quad 1.0 \exp \big(-5.0 \left(\mathbf{p}^{o_t}_z - \mathbf{p}^{b_t}_z \right) \big)\,+ \\
    &\quad 1.0\exp \big(-0.75 \left \|\Delta\theta(\mathbf{d}_{o_t}, \mathbf{d}_{b_t}) \right \| \big),\ \text{otherwise}
    \end{aligned}
    \right.
\end{equation}
where $\mathbf{p}^{o_t}$ and $\mathbf{p}^{b_t}$ represent the 3D positions of the object (chair/bed) surface center and the robot base in the world frame, respectively, while $\mathbf{d}_{o_t}$ and $\mathbf{d}_{b_t}$ denote the 2D horizontal unit vectors of the object and the robot base orientations in the world frame. The total reward function for \textit{Sit Down} is then given by:
\begin{equation}
r_t^{G\_sitDown} = r_t^{loco} + r_t^{sit}.
\end{equation}
For \textit{Lie Down}, once the robot has successfully sat on the bed, we introduce an additional reward to encourage lying down while facing upward toward the sky, combined with $r_t^{sit}$ and defined as:
\begin{equation}
r_t^{lie}\!=\!\left\{
    \begin{aligned}
    &r_t^{sit}, \  \left \| \mathbf{p}^{o_t}_{xy} - \mathbf{p}^{b_t}_{xy} \right \| < 0.3\ \text{and}\, \left( \mathbf{p}^{o_t}_{z} - \mathbf{p}^{b_t}_{z} \right) < 0.05\ \\
    &3.0+0.5 \exp \big(-0.75 \left \|\mathbf{D}_{world\_z}\cdot \mathbf{D}_{b_t}) \right \| \big)\,+ \\
    &\quad 0.5\exp \big(-2 \left \|\Delta\theta(\mathbf{d}_{o_t}^{\perp}, \mathbf{d}_{t}^{\dagger}) \right \| \big),\ \text{otherwise}
    \end{aligned}
    \right.
\end{equation}
Here, $\mathbf{D}_{world\_z}\in\mathbb{R}^3$ denotes the global vertical direction $[0,0,1]$, and $\mathbf{D}_{b_t}\in\mathbb{R}^3$ denotes the robot’s upward direction vector. The first term encourages the robot to face upward toward the sky. $\mathbf{d}_{o_t}^{\perp}$ is the 2D unit vector perpendicular to $\mathbf{d}_{o_t}$, and $\mathbf{d}_{t}^{\dagger}$ is the horizontal unit vector pointing from the head to the robot base. The second term encourages the robot to align parallel with the bed edge. Then the total reward function for \textit{Sit Down} is then given by:
\begin{equation}
r_t^{G\_lieDown} = r_t^{loco} + r_t^{lie}.
\end{equation}

\noindent \textbf{Scene Randomization} We randomize the task scene along the following three dimensions:
\begin{itemize}
\item The 2D position of the chair or bed is uniformly sampled within $[-5.0, 5.0]\,\mathrm{m}$ relative to the robot’s initial base position.
\item The height of both the chair and the bed is uniformly sampled from $[0.2, 0.5]\,\mathrm{m}$ above the ground.
\item The size of the chair and bed is randomized: the length and width of the chair are uniformly sampled from $[0.3, 0.6]\,\mathrm{m}$, while the bed length is uniformly sampled from $[1.2, 3.2]\,\mathrm{m}$ and its width from $[0.38, 0.63]\,\mathrm{m}$.
\end{itemize}

\subsubsection{\textbf{Stand Up}}
The humanoid starts in a seated position on the fixed chair and is tasked with standing up and walking toward a designated target location. Similarly, a thin platform is used to support stable sitting behaviors in simulation.

\noindent \textbf{Reference Motion Dataset} For locomotion, we use the same \textit{Loco} dataset as in the \textit{Carry Box} task. For the \textit{Stand Up} task, we additionally select two getting up sequences from SAMP.

In addition, to ensure initialization stability, we pre-collected a set of stable sitting poses on chairs of different heights using the sitting policy. These poses were used for initialization during training.

\noindent \textbf{Task Observations} The task-specific observation $\mathbf{o}_t^G\in\mathbb{R}^{12}$ comprises the following components:
\begin{itemize}[leftmargin=4mm]
\item Chair position $\mathbf{p}_{b_t}^{o_t} \in \mathbb{R}^3$
\item Chair rotation $\mathbf{R}_{b_t}^{o_t} \in \mathbb{R}^6$
\item Target position $\mathbf{p}_{b_t}^{g_t} \in \mathbb{R}^3$
\end{itemize}

\noindent \textbf{Task Rewards} In the first stage, the robot is encouraged to stand up to a target height, defined as:
\begin{equation}
r_t^{standup} = \left\{
    \begin{aligned}
    &3.0, \  \mathbf{p}^{b_t}_z >0.72 \\
    &3.0 \exp \big (-5.0 \left( 0.72 - \mathbf{p}^{b_t}_z \right) \big ).\  \text{otherwise}
    \end{aligned}
    \right.
\end{equation}
Once the robot has reached the target height, it is encouraged to walk toward the goal position, defined as:
\begin{equation}
    \begin{aligned}
    r_t^{loco\_tar} =\ &0.5 \exp \big (-5.0 \left \| 0.85 - \mathbf{d}_t^\prime \cdot \dot{\mathbf{p}}^{b_t}_{xy} \right \|^2 \big )\,+ \\
    &\quad\quad 0.5 \exp \big(-0.75 \left \|\Delta\theta(\mathbf{d}_t^\prime, \mathbf{d}_{b_t}) \right \| \big),
    \end{aligned}
\end{equation}
where $\mathbf{d}_t^\prime$ is a horizontal unit vector pointing from $\mathbf{p}_{xy}^{b_t}$ to $\mathbf{p}_{xy}^{g_t}$. The total reward function
for \textit{Stand Up} is then given by:
\begin{equation}
r_t^{G\_standUp} = r_t^{standup} + r_t^{loco\_tar}.
\end{equation}

\noindent \textbf{Scene Randomization} We randomize the task scene along the following three dimensions:
\begin{itemize}
\item The 2D position of the target is uniformly sampled within $[-5.0, 5.0]\,\mathrm{m}$ relative to the robot’s initial base position.
\item The height of the chair is uniformly sampled from $[0.2, 0.6]\,\mathrm{m}$ above the ground.
\item The length and width of the chair is uniformly sampled from $[0.38, 0.63]\,\mathrm{m}$.
\end{itemize}

\noindent \textbf{Configuration} In the original setting, we use a standing pose as the default pose. When initializing the robot from a seated pose, the policy often causes an abrupt upward jerk to transition into standing, disrupting training and causing instability. Therefore, for the \textit{Stand Up}, we set the default pose to a predefined seated pose, detailed in Table~\ref{tab:default_pos_standup}, to ensure a stable initial state.

\begin{table}[!h]
    \centering
    \vspace{-0.05in}
    \caption{Default Pose Configuration for the \textit{Stand Up} Task}
    \vspace{-0.05in}
    \begin{tabular}{ll!{\vrule width 0.5pt}ll}
    \toprule[1.0pt]
    \textbf{Term} & \textbf{Value} & \textbf{Term} & \textbf{Value} \\
    \midrule[0.5pt]
    left hip pitch joint  & $-1.2$ & right hip pitch joint & $-1.2$ \\
    left hip roll joint & $0.2$ & right hip yaw joint & $-0.2$ \\
    left hip yaw joint  & $0.0$ & right hip yaw joint  & $0.0$ \\ 
    left knee joint  & $1.2$ & right knee joint & $1.2$ \\
    left ankle pitch joint & $0.0$ & right ankle pitch joint & $0.0$ \\
    left ankle roll joint & $0.0$ & right ankle roll joint & $0.0$ \\
    
    left shoulder pitch joint & $0.2$ & right shoulder pitch joint & $0.2$ \\
    left shoulder roll joint & $0.8$ & right shoulder roll joint & $-0.8$ \\
    left shoulder yaw joint & $-0.7$ & right shoulder yaw joint & $0.7$ \\
    left elbow joint & $-0.3$ & right elbow joint & $-0.3$ \\
    left wrist roll joint & $0.0$ & right wrist roll joint & $0.0$ \\
    left wrist pitch joint & $0.0$ & right wrist pitch joint & $0.0$\\
    left wrist yaw joint & $0.0$ & right wrist yaw joint & $0.0$ \\

    waist yaw joint & $0.0$ & waist roll joint & $0.0$ \\
    waist pitch joint & $0.6$ \\
    
    \bottomrule[1.0pt]
    \vspace{-0.25in}
    \end{tabular}
    \label{tab:default_pos_standup}
\end{table}

\subsubsection{\textbf{Stylized Locomotion}}
The humanoid is tasked with tracking the given command $\mathbf{c}_t=[\mathbf{v}_x^c, \mathbf{v}_y^c, \bm{\omega}_\mathrm{yaw}^c] \in \mathbb{R}^3$ (denote the linear velocities in the longitudinal and lateral directions, and the angular velocity in the horizontal plane, respectively) while walking with a stylized gait, such as dinosaur-like walking or high-knee stepping.

\noindent \textbf{Reference Motion Dataset} We select two motion styles, \textit{Dinosaur} and \textit{HighKnees}, from the 100STYLE dataset~\cite{mason2022real}. Each style includes three sequences: forward walking, backward walking, and sidestep walking.

\noindent \textbf{Task Rewards} The only task reward for \textit{Stylized Locomotion} is to track the given linear and angular velocities, defined as:
\begin{equation}
    \begin{aligned}
    r_t^{G\_styleLoco}=\ &1.0 \exp \big(-4 \left \|\mathbf{v}_{xy}-\mathbf{v}_{xy}^c \right \|^2 \big)\ + \\
    &\quad \quad0.5\exp \big(-4 \left( \bm{\omega}_\mathrm{yaw}-\bm{\omega}_\mathrm{yaw}^c \right)^2 \big).
    \end{aligned}
\end{equation}

\subsection{Training Details}
\subsubsection{Regularization Rewards}
The regularization reward $r_t^R$ is summarized in Table~\ref{tab:regu_reward}.
\begin{table}[!h]
    \centering
    \vspace{-0.05in}
    \caption{Regularization Reward Functions}
    \vspace{-0.05in}
    \begin{tabular}{ll!{\vrule width 0.5pt}ll}
    \toprule[1.0pt]
    \textbf{Term} & \textbf{Weight} & \textbf{Term} & \textbf{Weight} \\
    \midrule[0.5pt]
    dof velocity  & $-2e-4$ & torques  & $-1e-4$\\ [0.2ex]
    dof acceleration  & $-1e-7$ & torque limits  & $-0.1$\\
    dof position limits  & $-5.0$ & action rate & $-0.03$\\ [0.2ex]
    dof velocity limits  & $-1e-3$ &  & \\
    \bottomrule[1.0pt]
    \vspace{-0.25in}
    \end{tabular}
    \label{tab:regu_reward}
\end{table}

\subsubsection{Domain Randomization}
To enhance robustness and facilitate sim-to-real transfer, we employ domain randomization, summarized in Table~\ref{tab:domain_random}.

\begin{table}[!h]
    \centering
    \vspace{-0.05in}
    \caption{Domain Randomization Settings}
    \vspace{-0.05in}
    \begin{tabular}{ll}
    \toprule[1.0pt]
    \textbf{Term} & \textbf{Value}\\
    
    \midrule[0.8pt]
    \multicolumn{2}{c}{\textbf{Observations}} \\ [0.3ex]
    angular velocity noise & $\mathcal{U}(-0.3, 0.3)$ rad/s \\ [0.1ex]
    joint position noise & $\mathcal{U}(-0.02, 0.02)$ rad/s \\ [0.1ex]
    joint velocity noise & $\mathcal{U}(-2.0, 2.0)$ rad/s \\ [0.1ex]
    projected gravity noise & $\mathcal{U}(-0.05, 0.05)$ rad/s \\ [0.1ex]
    FK noise & $\mathcal{U}(-0.05, 0.05)$ m \\
    
    \midrule[0.5pt] 
    \multicolumn{2}{c}{\textbf{Humanoid Physical Properties}} \\ [0.3ex]
    actuator offset & $\mathcal{U}(-0.05, 0.05)$ rad \\ [0.1ex]
    motor strength noise & $\mathcal{U}(0.9, 1.1)$ \\ [0.1ex]
    payload mass & $\mathcal{U}(-2.0, 2.0)$ kg \\ [0.1ex]
    center of mass displacement & $\mathcal{U}(-0.05, 0.05)$ m \\ [0.1ex]
    Kp, Kd noise factor & $\mathcal{U}(0.85, 1.15)$ \\ [0.1ex]

    \midrule[0.5pt] 
    \multicolumn{2}{c}{\textbf{Object Dynamics}} \\ [0.3ex]
    box friction factor & $\mathcal{U}(0.5, 1.2)$ \\ [0.1ex]
    box restitution factor & $\mathcal{U}(0.0, 0.2)$ \\ [0.1ex]
    platform friction factor & $\mathcal{U}(0.5, 1.2)$ \\ [0.1ex]

    \midrule[0.5pt] 
    \multicolumn{2}{c}{\textbf{Object Localization}} \\ [0.3ex]
    position offset & $\mathcal{U}(-0.05, 0.05)$ m \\ [0.1ex]
    position noise & $\mathcal{U}(-0.05, 0.05)$ m \\ [0.1ex]
    rotation offset & $\mathcal{U}(-5.0, 5.0)^\circ$ \\ [0.1ex]
    rotation noise & $\mathcal{U}(-5.0, 5.0)^\circ$ \\ [0.1ex]
    \bottomrule[1.0pt]
    \end{tabular}
    \label{tab:domain_random}
\end{table}

\subsubsection{Hyperparameters}
The hyperparameters used for training is summarized in Table~\ref{tab:hyperparameters}.
\begin{table}[!h]
    \centering
    \vspace{-0.05in}
    \caption{Hyperparameters}
    \vspace{-0.05in}
    \begin{tabular}{ll}
    \toprule[1.0pt]
    \textbf{Hyperparameter} & \textbf{Value} \\

    \midrule[0.8pt] 
    \multicolumn{2}{c}{\textbf{General}} \\
    num of robots & 4096 \\
    num of steps per iteration & 100 \\
    num of epochs & 5 \\
    gradient clipping & 1.0 \\
    adam epsilon & $1e-8$ \\

    \midrule[0.5pt]
    \multicolumn{2}{c}{\textbf{PPO}} \\
    clip range & 0.2 \\
    entropy coefficient & 0.01 \\
    discount factor $\gamma$ & 0.99 \\
    GAE balancing factor $\lambda$ & 0.95 \\
    desired KL-divergence & 0.01 \\
    actor and double critic NN & MLP, hidden units [512, 256, 256] \\

    \midrule[0.5pt]
    \multicolumn{2}{c}{\ours} \\
    reward coefficient $w^S$ (general) & 0.3 \\
    reward coefficient $w^G$, $w^R$ & 0.7, 0.7 \\
    gradient penalty $w^\mathrm{gp}$ & 1.0 \\
    distance threshold $\epsilon$ & 0.6 \\
    AMP discriminator NN & MLP, hidden units [512, 256, 256] \\
    
    \bottomrule[1.0pt]
    \end{tabular}
    \label{tab:hyperparameters}
\end{table}

\subsection{Deployment Details}

\begin{figure}[!h]
    \centering
    \includegraphics[width=0.85\linewidth]{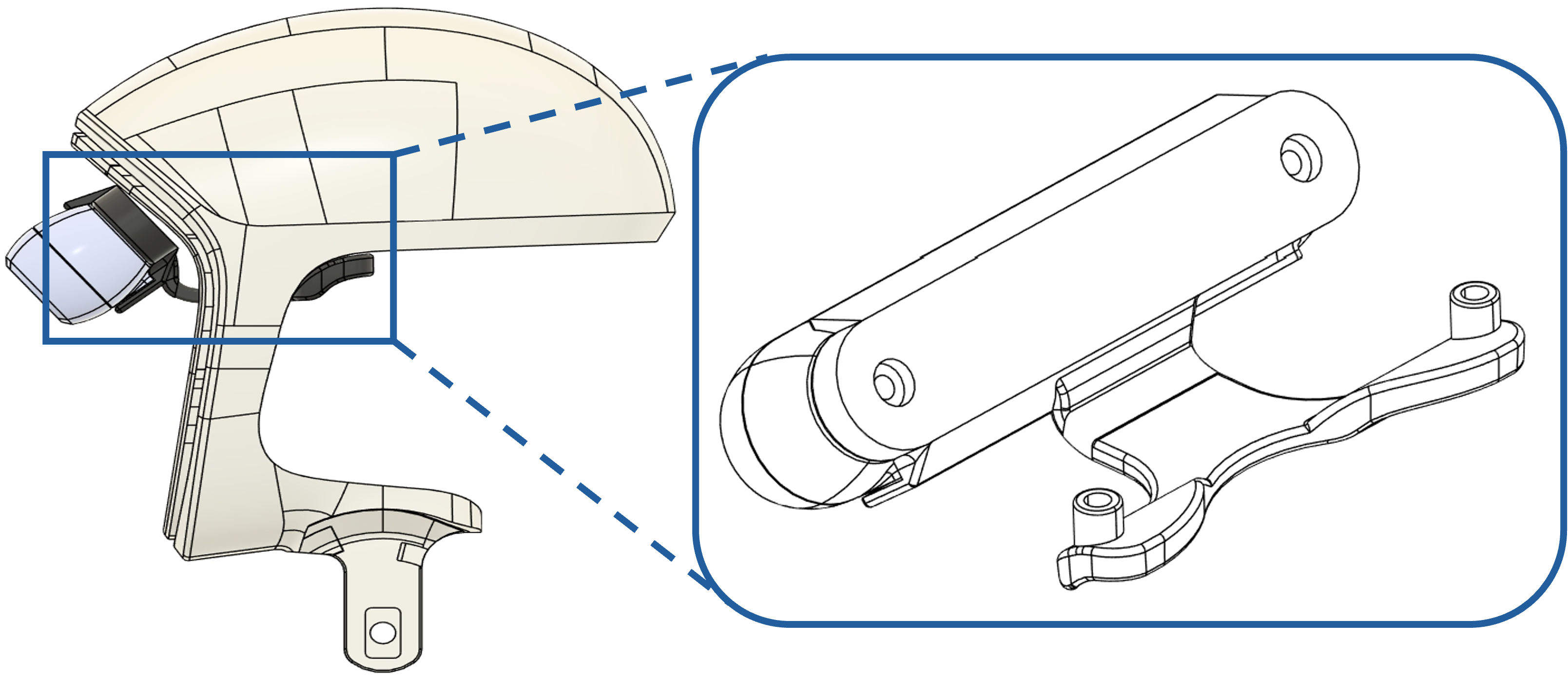}
    \caption{3D model of the D455 camera bracket.
    \vspace{-0.15in}
    }
    \label{fig:hardware}
\end{figure}

To support the Intel RealSense D455 camera mounted on the humanoid’s head, we designed a 3D-printed camera bracket. The bracket is fixed to the torso link with an offset of $(0.08,0.01,0.40)\,\mathrm{m}$ and rotated by approximately $40^\circ$ about the pitch axis, as shown in Fig.~\ref{fig:hardware}.

\subsection{Evaluation Details}

\subsubsection{Baseline Implementation Details}

For \textbf{RL-Rewards}, we replace the style reward $r_t^S$ with explored RL-based gait rewards~\cite{long2025learning, wang2025beamdojo} to regularize the gait during task execution. The total reward function is formulated as 
\begin{equation}
    r_t^{RL} = r_t^{Gait} + r_t^{R} + r_t^{G},
\end{equation}
where $r_t^{R}$ and $r_t^{G}$ are the same as in \ours. The gait reward terms are summarized in Table~\ref{tab:gait_reward}.

\begin{table}[!h]
    \centering
    \vspace{-0.05in}
    \caption{Gait Reward Functions}
    \vspace{-0.05in}
    \begin{tabular}{ll!{\vrule width 0.5pt}ll}
    \toprule[1.0pt]
    \textbf{Term} & \textbf{Weight} & \textbf{Term} & \textbf{Weight} \\
    \midrule[0.5pt]
    base height  & $-10.0$ & feet clearance & $-0.5$ \\ [0.2ex]
    z velocity  & $-2.0$ & feet air time & $0.05$ \\
    roll-pitch velocity  & $-0.05$ & feet distance  & $0.5$ \\ [0.2ex]
    orientation  & $-1.0$ & hip joint deviation & $-0.5$ \\
    \bottomrule[1.0pt]
    \vspace{-0.25in}
    \end{tabular}
    \label{tab:gait_reward}
\end{table}

For \textbf{Tracking-Based}, we employ the tracking reward $r_t^{Track}$, primarily adapted from~\cite{he2025asap}, together with the regularization reward $r_t^{R}$. The tracking reward terms are summarized in Table~\ref{tab:track_reward}.

\begin{table}[!h]
    \centering
    \vspace{-0.05in}
    \caption{Tracking Reward Functions}
    \vspace{-0.05in}
    \begin{tabular}{ll!{\vrule width 0.5pt}ll}
    \toprule[1.0pt]
    \textbf{Term} & \textbf{Weight} & \textbf{Term} & \textbf{Weight} \\
    \midrule[0.5pt]
    body position  & $1.0$ & base linear velocity & $0.5$ \\ [0.2ex]
    body rotation  & $0.5$ & end-effector position & $1.5$ \\
    base position  & $1.0$ & joint position  & $1.0$ \\ [0.2ex]
    base rotation  & $0.5$ & object position (\textit{Carry Box}) & $1.0$ \\
    \bottomrule[1.0pt]
    \vspace{-0.25in}
    \end{tabular}
    \label{tab:track_reward}
\end{table}

\subsubsection{Success Rate Evaluation}

The success criteria for each task are defined as follows:
\begin{itemize}[leftmargin=4mm]
\item \textit{Carry Box}: The box is placed at the target position with a distance error of less than $0.1\,\mathrm{m}$.
\item \textit{Sit Down} and \textit{Lie Down}: The robot’s base position is within $0.1\,\mathrm{m}$ of the target location, with its heading aligned to the chair/bed orientation. For \textit{Lie Down}, the body direction from head to feet must also be parallel to the bed, both with a tolerance of $15^\circ$.
\item \textit{Stand Up}: The robot successfully stands up from the chair with a base height exceeding $0.72\,\mathrm{m}$, and then reaches the target position with a distance error of less than $0.3\,\mathrm{m}$.
\end{itemize}

\subsubsection{Human-Likeness Score Evaluation}

We evaluate the human-likeness score $S_\mathrm{human}$ using Gemini-2.5-Pro~\cite{comanici2025gemini}. The model is prompted with task descriptions and experimental trajectories, and outputs a human-likeness score in the range $[0, 5]$ for each demonstration clip. Below is an example prompt used for evaluating the \textit{Lie Down} task:

\begin{lstlisting}[language={}]
I will provide you with three videos of a robot performing a lie-down task, named LieDown_PhysHSI_1, LieDown_Mimic_1, and LieDown_RL_1.

Please analyze and compare them based on the criterion of Naturalness. For this task, Naturalness is defined by how closely the robot's movement resembles a natural, human-like action. When evaluating, please consider these specific aspects:

1. Fluidity and Smoothness: Is the motion continuous, or is it jerky and segmented?
2. Stability and Balance: Does the robot appear stable and in control, or does it look wobbly and at risk of falling?
3. Plausibility of Strategy: Does the robot use its limbs and body in a way a human would (e.g., using hands for support, bending knees, controlled descent)?

Please provide a Naturalness score out of 5 (decimals are allowed) for each video in a table. After the scores, write a summary explaining the key differences and justifying your ratings based on the aspects mentioned above.
\end{lstlisting}

\end{document}